\documentclass[10pt,twocolumn,letterpaper]{article}

 \usepackage{cvpr}              
\usepackage{multirow}
\usepackage{amssymb}
\usepackage{booktabs}
\usepackage{multirow}
\usepackage{siunitx}     
\usepackage{pifont}       
\usepackage{xcolor}
\usepackage{booktabs}
\usepackage{siunitx}
\usepackage[accsupp]{axessibility}
\definecolor{cvprblue}{rgb}{0.21,0.49,0.74}
\usepackage[pagebackref,breaklinks,colorlinks,allcolors=cvprblue]{hyperref}
\def\paperID{15963} 
\def\confName{CVPR}
\def\confYear{2026}

\title{From Contrast to Consistency: Rethinking Event-based \\ Continuous-Time Optical Flow Estimation}

\author{
Rui Hu$^{1}$\textsuperscript{*}\quad
Song Wu$^{1}$\textsuperscript{*}\quad
Wen Yang$^{1}$ \quad
Jinjian Wu$^{1}$\textsuperscript{\dag}\\
Xidian University$^{1}$\\
{\tt\small rui.hu@stu.xidian.edu.cn, swu\_666@stu.xidian.edu.cn}
}

\begin{document}
\maketitle
\maketitle

\maketitle

\def\thefootnote{}\footnotetext{\textsuperscript{*} Equal contribution. \textsuperscript{\dag} Corresponding author.}\def\thefootnote{\arabic{footnote}}

\begin{abstract}

Estimating continuous optical flow is a fundamental yet challenging problem in dynamic visual perception. Event-based cameras, with microsecond latency and high dynamic range, capture brightness changes asynchronously, offering a unique opportunity to model motion with fine temporal precision. However, the scarcity of temporally dense ground-truth annotations limits the effectiveness of supervised learning, while contrast maximization (CM) frameworks, focused on sharpening the Image of Warped Events (IWE), often neglect temporal continuity and structural coherence, leading to distorted trajectories under complex motion.
To overcome these challenges, we propose a hybrid-supervised framework for continuous-time optical flow estimation, grounded in the principle of Spatio-temporal Structural Consistency (STSC). This paradigm jointly enforces local structural stability and trajectory continuity, ensuring physically coherent motion across time. To further enhance representation and robustness, we design a bidirectionally complementary multi-scale architecture and employ a curriculum-guided hybrid training strategy, enabling a smooth transition from supervised point constraints to self-supervised manifold regularization.
Comprehensive experiments across multiple benchmarks show that our method achieves state-of-the-art performance in both continuous-time and standard optical flow estimation, demonstrating the effectiveness of the proposed learning paradigm.

\end{abstract}

\section{Introduction}

\begin{figure}[htbp]
    \centering
    \includegraphics[width=1\linewidth]{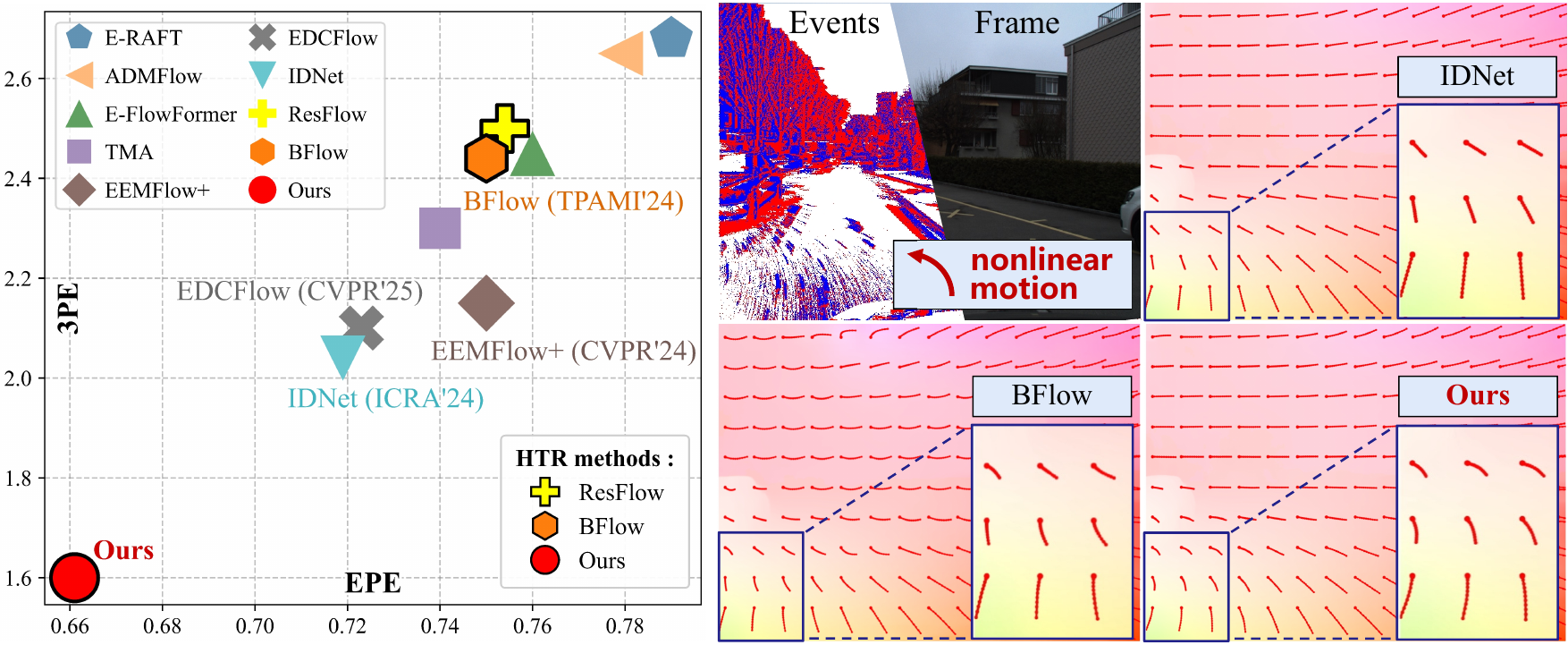}
    \caption{Left: State-of-the-art (SOTA) comparison on the DSEC-Flow benchmark~\cite{gehrig2021dsec}. Our method achieves the best performance (bottom-left). Right: Qualitative visualization in a non-linear motion scene. Compared to the IDNet~\cite{wu2024lightweight} and the leading continuous-time method BFlow~\cite{gehrig2024dense}, only our approach produces continuous flow results that are physically consistent with the true motion.}
    \label{fig:motivation}
\end{figure}

Optical flow estimation remains the most prevalent approach for modeling motion dynamics and has been widely applied across diverse computer vision tasks, including deblurring~\cite{lin2022flow,zhang2022spatio}, frame interpolation~\cite{kim2023event,tulyakov2021time,tulyakov2022time}, and object tracking~\cite{yao2023folt,vserych2023planar}. Despite its ubiquity, conventional imaging is inherently constrained by discrete exposure intervals, rendering it ill-suited for capturing the rapid, nonlinear motion patterns that characterize real-world dynamics.

Event cameras~\cite{gallego2020event} offer a paradigm shift in visual sensing. Characterized by microsecond latency and high dynamic range, they asynchronously activate individual pixels in response to intensity fluctuations.  This distinctive sensing mechanism provides the raw ingredients for continuous-time optical flow estimation at high temporal resolution (HTR), where fine-grained motion cues are indispensable. 

However, event-based continuous-time optical flow estimation remains a formidable challenge. The major bottleneck is the absence of temporally dense ground truth, which prevents supervised learning from fully exploiting the temporal fidelity of event data. Moreover, the intrinsic sparsity and nonlinear temporal distribution of events further complicate optimization, making it difficult to learn stable and coherent motion representations.

Recent approaches~\cite{hamann2024motion,paredes2023taming} attempt to alleviate these limitations by optimizing self-supervised objectives within the contrast maximization (CM) framework~\cite{gallego2018unifying}. However, since this paradigm centers on enhancing the sharpness of the Image of Warped Events (IWE), it inherently overlooks the temporal continuity and structural coherence of motion trajectories, making it difficult to model continuously evolving motion under complex or nonlinear dynamics.

In contrast to prior CM-based paradigms, we observe that events triggered by the same physical surface inherently preserve local structural patterns as they evolve through motion, forming a temporally stable spatio-temporal manifold. Motivated by this insight, we revisit the problem from a new perspective: rather than treating events as unordered points awaiting alignment, we interpret them as samples drawn from an intrinsically structured spatio-temporal manifold that encodes the continuity of physical motion. Building upon this interpretation, we introduce the principle of Spatio-temporal Structural Consistency (STSC) and establish a novel hybrid-supervised optimization paradigm that constrains learning from two complementary aspects—local structural stability and trajectory continuity. These constraints guide the network to reconstruct the true physical motion field, rather than merely optimizing endpoint alignment for contrast enhancement.

Designing an effective architecture for continuous-time motion estimation requires balancing spatial representation and temporal precision. To this end, we construct a bidirectional complementary multi-scale architecture that encodes motion cues at multiple spatial scales while employing a bidirectional refinement update module to maintain temporal consistency and handle occlusions and nonlinear motion.  Moreover, given the scarcity of temporally dense ground-truth annotations, neither purely supervised nor self-supervised schemes can sufficiently capture continuous spatio-temporal dynamics. We therefore propose a curriculum-guided anchored hybrid training strategy that progressively transitions from supervised point-wise constraints to self-supervised STSC regularization, enabling the network to learn a temporally coherent and physically consistent motion field.

With these components, our method enables supervised training at 10 Hz while theoretically achieving unbounded inference temporal resolution. Extensive evaluations across multiple benchmarks demonstrate that our approach significantly outperforms existing state-of-the-art methods. Representative results are illustrated in Fig.~\ref{fig:motivation}.

Our contributions can be summarized as:

\begin{itemize}
\item We propose a novel hybrid-supervised paradigm for continuous optical flow estimation grounded in the spatio-temporal structural consistency of events, enabling the recovery of motion fields that more faithfully reflect real physical dynamics.
\item We design a spatially efficient and temporally complementary architecture that captures complex spatio-temporal dynamics through multi-scale spatial encoding and bidirectional temporal refinement.
\item We introduce a curriculum-guided hybrid training strategy that smoothly transitions from supervised point-wise constraints to self-supervised manifold regularization, ensuring stable and progressive learning.
\item We conduct comprehensive evaluations across multiple datasets and tasks, demonstrating that our method consistently achieves state-of-the-art performance.
\end{itemize}

\begin{figure*}[h]
    \centering
    \includegraphics[width=1\linewidth]{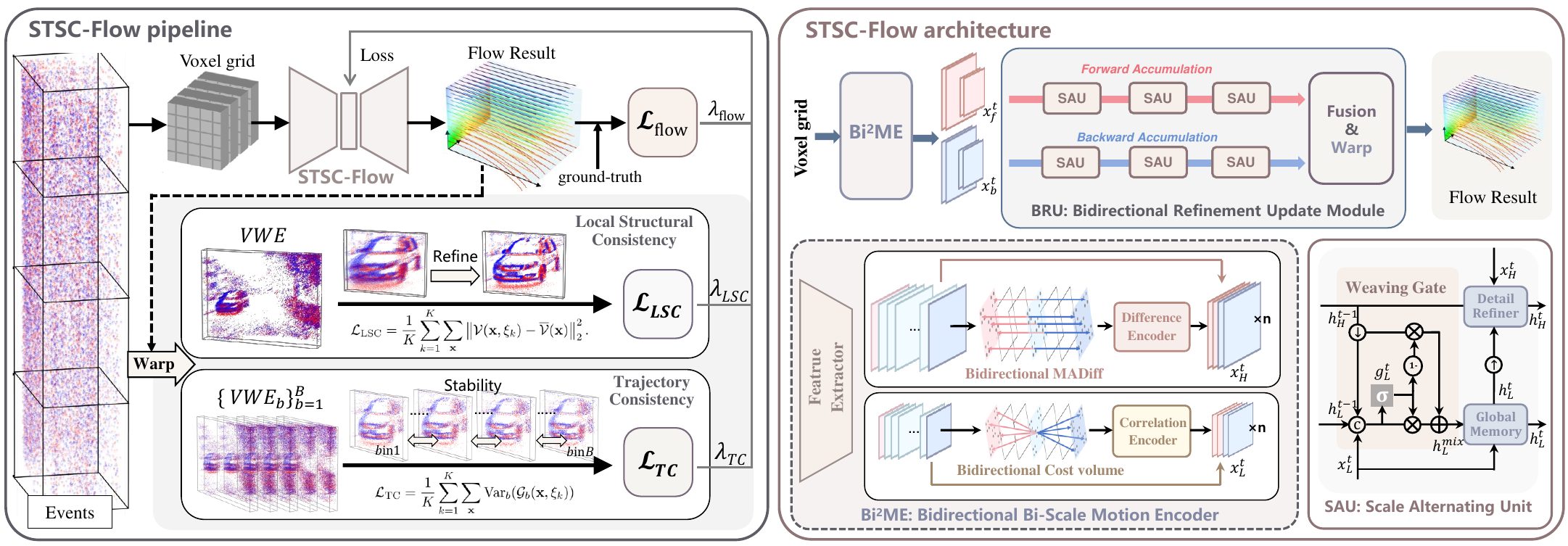}
    \caption{Overall framework of the proposed STSC-Flow, detailing the training pipeline (Left) and the network architecture (Right).
    \textbf{Left: The hybrid supervision pipeline.} The network (right) predicts a continuous flow, which is constrained by two objectives: (1) a supervised endpoint loss $\mathcal{L}_{\text{flow}}$ against sparse ground truth, and (2) our STSC self-supervision ($\mathcal{L}_{\mathrm{LSC}}, \mathcal{L}_{\mathrm{TC}}$), computed on the Volumetric Warped Events (VWE). A curriculum learning strategy (Sec.~\ref{sec:CL}) balances these losses during training.
    \textbf{Right: The network architecture.} (1) \textbf{Bi$^2$ME} extracts bidirectional, multi-scale motion features from the input voxel grids. (2) The \textbf{BRU} module, which embeds \textbf{SAUs} for cross-scale fusion, performs symmetric forward and backward temporal accumulation. (3) The final continuous flow estimate is produced by aggregating the bidirectional states.}
    \label{fig:architecture}
\end{figure*}

\section{Related Work}
\label{sec:related_work}

\subsection{Event-based Optical Flow Estimation}
Early event-based optical flow research \cite{benosman2013event, mueggler2015lifetime, bardow2016simultaneous} adapted classical algorithms, such as Lucas-Kanade \cite{lucas1981iterative}, with handcrafted heuristics. However, these methods were limited to sparse or uniform motion settings, prompting the field's shift toward learning-based methods, which have now become the dominant paradigm.
Among these, a seminal line of work, inspired by RAFT \cite{teed2020raft}, has demonstrated state-of-the-art performance. E-RAFT \cite{gehrig2021raft} pioneered this adaptation, introducing event voxel grids to preserve fine-grained spatio-temporal structure and combining them with 4D correlation volumes for iterative refinement. Subsequent methods \cite{liu2023tma, wan2022learning, li2023blinkflow, gehrig2024dense} have further refined this RAFT-inspired paradigm. However, the high cost of 4D correlation volumes typically restricts correspondence matching to a coarse resolution. This bottleneck prompted alternative designs like IDNet \cite{wu2024lightweight} and EDCFlow \cite{liu2025edcflow}, which bypass the costly correlation step using iterative deblurring or differential operators.


\subsection{Event-based Continuous-Time Flow Estimation}
\label{sec:related_work_continuous}

The high temporal resolution of the event data enables continuous-time optical flow estimation. However, this field is severely constrained by a fundamental problem: the lack of dense HTR ground truth in real-world datasets. To mitigate this supervision gap, existing methods apply self-supervised constraints to the intermediate motion, typically via contrast maximization(CM) over warped events \cite{gallego2018unifying}. However, these contrast-based losses, lacking ground-truth anchoring, suffer from sub-optimal performance and are susceptible to ``Projection Collapse'' \cite{shiba2022secrets}. Another category attempts to use sparse GT for implicit supervision of HTR flow, often through accumulation-based schemes such as EVA-Flow \cite{ye2025towards} or residual refinement as in ResFlow \cite{zhou2025resflow}. However, the optimization objective for such implicit supervision is ambiguous, failing to guarantee the physical realism of the intermediate trajectories.

A more explicit approach is to directly parameterize the complete motion trajectory. BFlow \cite{gehrig2024dense} follows this direction by representing pixel paths with Bézier curves and predicting their control points. Nevertheless, on real-world datasets, this method relies solely on sparse LTR-GT (i.e., trajectory endpoints), which provides neither sufficient constraints for complex non-linear motion nor adequate motion priors to regularize the continuous trajectory. 
In contrast, our method introduces a principled self-supervision mechanism for continuous-time flow through spatio-temporal Structural Consistency (STSC). By jointly enforcing local structural stability and temporal trajectory coherence, STSC provides the dense motion priors missing in previous contrast-based or implicitly supervised approaches. 


\section{Method}

Our method is built upon the principle of Spatio-temporal Structural Consistency (STSC), which reformulates the traditional contrast-based self-supervision paradigm into a consistency-aware learning framework.
To achieve this, we first introduce the definition of spatio-temporal structural consistency and its corresponding self-supervised objectives (Sec.~\ref{sec:esc}).
We then describe the proposed network architecture, which combines bidirectional temporal refinement with multi-scale spatial encoding to capture complex spatio-temporal motion patterns (Sec.~\ref{sec:arc}).
Finally, we detail the curriculum-guided hybrid training strategy, enabling a smooth transition from supervised point-wise constraints to self-supervised manifold regularization for stable and continuous learning (Sec.~\ref{sec:CL}). The overall framework is illustrated in Fig.~\ref{fig:architecture}.

\subsection{Spatial-temporal Structural Consistency}
\label{sec:esc}

We begin our formulation from a fundamental observation: objects undergoing continuous motion generate events that lie on a smooth spatio-temporal manifold, reflecting the continuity of the underlying physical process.
Temporally, events triggered by the same physical point should follow coherent trajectories; 
spatially, the local edge or surface structure should remain stable over short temporal intervals. 
To encode these physical priors, we formulate \emph{Spatio-temporal Structural Consistency (STSC)} as the joint preservation of local spatial structure and temporal coherence.

\paragraph{Volumetric Warped Events.}

Conventional self-supervised methods~\cite{gallego2018unifying} typically warp all events to a single reference time, which removes temporal continuity and may introduce structural distortion.
Instead, we align each source-time bin to a common reference center while preserving its within-bin relative-time structure, yielding a spatio-temporal volume in a shared reference frame.

\begin{figure}[t]
    \centering
    \includegraphics[width=1\linewidth]{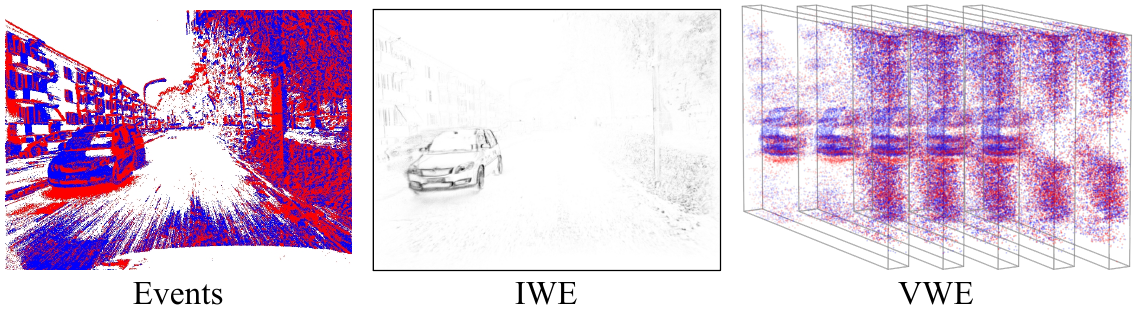}
    \caption{Comparison of IWE and VWE. \textbf{Left:} input events. \textbf{Middle:} IWE projects warped events onto a single time slice. 
    \textbf{Right:} VWE represents aligned events as a 3D spatio-temporal volume while preserving their relative-time structure.}
    \label{fig:vwe}
\end{figure}

Let $\mathcal{E}=\{(\mathbf{x}_i,t_i,\sigma_i)\}_{i=1}^{N}$ denote the input event stream, and let $\{\mathcal{I}_b\}_{b=1}^{B}$ be $B$ temporal bins with centers $\{c_b\}_{b=1}^{B}$.
Given a reference time $t_0$, we define the bin-wise temporal shift as $\Delta_b=t_0-c_b$ and the preserved relative time as $\xi_i=t_i-c_b$ for each event in bin $\mathcal{I}_b$.
The bin-wise \emph{Volumetric Warped Events (VWE)} is then defined as
{\small
\begin{equation}
\mathrm{VWE}_b(\mathbf{x},\xi)
=
\sum_{i:\,t_i\in\mathcal{I}_b}
\sigma_i\,
\kappa_s\!\Big(
\mathbf{x}-\mathcal{W}_{t_i\rightarrow t_i+\Delta_b}(\mathbf{x}_i)
\Big)\,
\kappa_t\!\big(\xi-\xi_i\big).
\label{eq:new_vwe_bin}
\end{equation}
}
and obtain the full volume by
\begin{equation}
\mathrm{VWE}(\mathbf{x},\xi)
=
\sum_{b=1}^{B}\mathrm{VWE}_b(\mathbf{x},\xi).
\label{eq:new_vwe}
\end{equation}
Unlike single-time warping, the proposed VWE preserves both cross-bin alignment and within-bin temporal micro-structure (Fig.~\ref{fig:vwe}).

\paragraph{Local Structural Consistency.}
After alignment, the aggregated volume should exhibit stable local patterns along the relative-time axis, rather than fluctuating across phases due to inaccurate motion compensation.
To capture this property, we directly regularize the full VWE by encouraging each relative-time phase to remain close to the mean local structure of the entire volume.
For each relative time $\xi$, we first aggregate a local temporal neighborhood:
\begin{equation}
\mathcal{V}(\mathbf{x},\xi)
=
\int_{\xi-r}^{\xi+r}
w_{\xi}(s-\xi)\,
\mathrm{VWE}(\mathbf{x},s)\,ds,
\label{eq:lsc_agg}
\end{equation}
where $w_{\xi}(\cdot)$ is a local weighting kernel on the relative-time axis.
Given $K$ discrete relative-time samples $\{\xi_k\}_{k=1}^{K}$, the mean local structure is defined as
\begin{equation}
\overline{\mathcal{V}}(\mathbf{x})
=
\frac{1}{K}
\sum_{k=1}^{K}
\mathcal{V}(\mathbf{x},\xi_k).
\label{eq:lsc_mean}
\end{equation}
We then enforce local structural consistency by penalizing the deviation of each relative-time phase from this mean structure:
\begin{equation}
\mathcal{L}_{\mathrm{LSC}}
=
\frac{1}{K}
\sum_{k=1}^{K}
\sum_{\mathbf{x}}
\left\|
\mathcal{V}(\mathbf{x},\xi_k)-\overline{\mathcal{V}}(\mathbf{x})
\right\|_2^2.
\label{eq:new_lsc}
\end{equation}
By suppressing structural variation across relative-time phases, $\mathcal{L}_{\mathrm{LSC}}$ encourages the aligned event volume to preserve coherent local spatio-temporal organization, thereby yielding a more faithful reconstruction of the underlying motion manifold.

\paragraph{Trajectory Consistency.}

While $\mathcal{L}_{\mathrm{LSC}}$ stabilizes the aggregated volume, trajectory consistency further constrains whether aligned source bins follow coherent motion traces in the shared reference frame.
To characterize trajectory geometry, we compute the spatio-temporal gradient field of each normalized bin-wise volume:
\begin{equation}
\mathcal{G}_b(\mathbf{x},\xi)
=
\nabla_{(\mathbf{x},\xi)}
\,\mathrm{norm}\!\left(
\mathrm{VWE}_b(\mathbf{x},\xi)
\right),
\label{eq:tc_grad}
\end{equation}
where $\nabla_{(\mathbf{x},\xi)}$ denotes the joint gradient with respect to the spatial coordinates and the relative-time axis.
Given $K$ discrete relative-time samples $\{\xi_k\}_{k=1}^{K}$, trajectory consistency is enforced by directly minimizing the variance of these gradient fields across source bins:
\begin{equation}
\mathcal{L}_{\mathrm{TC}}
=
\frac{1}{K}
\sum_{k=1}^{K}
\sum_{\mathbf{x}}
\operatorname{Var}_{b}
\!\left(
\mathcal{G}_b(\mathbf{x},\xi_k)
\right),
\label{eq:new_tc}
\end{equation}

Minimizing $\mathcal{L}_{\mathrm{TC}}$ constrains the motion field to produce temporally coherent trajectories, ensuring that events belonging to consecutive timestamps follow smooth and coherent motion trajectories in the spatio-temporal manifold.

\subsection{Architecture}
\label{sec:arc}

To implement the proposed Spatial-temporal Structure Consistency, 
we introduce the STSC-Flow architecture (Fig.~\ref{fig:architecture}) to estimate continuous-time optical flow. 
The input events are first converted into voxel grids, which are then processed by the 
\textbf{Bi$^2$ME} encoder to extract bidirectional multi-scale motion features. Subsequently, these features are refined by the \textbf{BRU} module, which performs symmetric forward–backward updates using \textbf{SAUs} to enforce cross-scale and temporal coherence. Finally, the resulting bidirectional flow predictions are aggregated to produce the final temporally coherent estimation.


\paragraph{Event Representation.}
We adopt the event voxel grid representation~\cite{gehrig2021raft,ye2025towards} by dividing the time window into $B$ temporal bins of width $\tau$. 
Events within each bin are aggregated into a voxel slice $\mathbf{V}_t(x,y)$ using trilinear interpolation:
{\small
\begin{equation}
\begin{aligned}
\mathbf{V}_t(x,y) &= \sum_i p_i \, k(x - x_i)\, k(y - y_i)\,k\left( \frac{t_i - t_t}{\tau} \right), \\
k(a) &= \max(0, 1 - |a|),
\end{aligned}
\end{equation}
}
where $t_t$ is the center timestamp of the $t$-th bin, and $k(a)$ is the 1D linear interpolation kernel. This process yields a sequence of temporally fine-grained voxel grids $\{\mathbf{V}_t\}_{t=1}^{B}$, which serve as input to the subsequent modules.

\paragraph{Bi²ME: Bi-Scale Bidirectional Motion Encoder.}

To handle spatially heterogeneous motion, we propose Bi$^2$ME, a dual-scale bidirectional encoder that captures both fine and global motion dynamics. It extracts motion features at two complementary resolutions: 
a low-resolution branch $\{F_L^t\}$ capturing global motion context, 
and a high-resolution branch $\{F_H^t\}$ preserving detailed structures. 
Bidirectional correlation volumes are constructed to anchor motion to the temporal boundaries:


\begin{equation}
C_f^t = \frac{F_L^1\,(F_L^t)^\top}{\sqrt{D}}, \quad C_b^t = \frac{F_L^t\,(F_L^B)^\top}{\sqrt{D}}.
\end{equation}
Here, $F_L^1, F_L^B$ are the features at the first and last bins, and $D$ is the feature dimension. 
To enhance sensitivity to the fine structures critical for \textit{Local Spatial Consistency}, we apply a \textit{Motion-aware Difference (MADiff)} operation on the high-resolution features:
\begin{equation}
M_{\text{f}}^t = F_H^t - F_H^1, \quad M_{\text{b}}^t = F_H^t - F_H^B.
\end{equation}
The correlation volumes ($C_f^t, C_b^t$) and the difference features ($M_f^t, M_b^t$) are fused with their respective feature streams ($F_L^t, F_H^t$) to produce the final bidirectional motion-enhanced feature sequence $\{({x}_{L,\mathrm{f}}^t, {x}_{L,\mathrm{b}}^t, {x}_{H,\mathrm{f}}^t, {x}_{H,\mathrm{b}}^t)\}_{t=1}^B$ as input to the refinement module.

\paragraph{SAU: Scale Alternating Unit.}
To couple coarse-scale temporal motion with fine-scale spatial details, 
we introduce the Scale Alternating Unit (SAU), a dual-branch recurrent structure. Each SAU contains a \textit{Global Memory Unit (GMU)} at $1/8$ resolution and a \textit{Detail Refiner Unit (DRU)} at $1/4$ resolution. 
A \textbf{Weaving Gate} adaptively mixes information between the two branches:
\begin{equation}
\begin{aligned}
g_L^t &= \sigma\left(\mathrm{Conv}_{3\times3}\left([h_L^{t-1}, \tilde{h}_H^{t-1}, x_L^t]\right)\right), \\
h_L^{\text{mix}} &= g_L^t \odot \tilde{h}_H^{t-1} + (1 - g_L^t) \odot h_L^{t-1}.
\end{aligned}
\end{equation}
GMU and DRU then update their hidden states using the mixed context and the current inputs. This cross-scale fusion strategy improves both spatial precision and temporal consistency.

\paragraph{BRU: Bidirectional Refinement Update.}
Unidirectional recurrent updates correspond to a first-order difference scheme, whereas combining past and future information yields a second-order central-difference approximation. We adopt this principle as the theoretical basis of our refinement design, with the full derivation provided in the \textit{supplementary material}. Accordingly, BRU employs two parallel SAUs that traverse the sequence in opposite directions to obtain temporally unbiased features.


\noindent\textbf{Forward Accumulation ($t = 1 \rightarrow B$):}
\begin{equation}
\begin{cases}
h_{L,\mathrm{f}}^t = \mathrm{GMU}(h_{L,\mathrm{f}}^{t-1},\, \tilde{h}_{H,\mathrm{f}}^{t-1},\, x_{L,\mathrm{f}}^t) \\[6pt]
h_{H,\mathrm{f}}^t = \mathrm{DRU}(h_{H,\mathrm{f}}^{t-1},\, x_{H,\mathrm{f}}^t,\, \tilde{h}_{L,\mathrm{f}}^t)
\end{cases}
\end{equation}

\noindent\textbf{Backward Accumulation ($t = B \rightarrow 1$):}
\begin{equation}
\begin{cases}
h_{L,\mathrm{b}}^t = \mathrm{GMU}(h_{L,\mathrm{b}}^{t+1},\, \tilde{h}_{H,\mathrm{b}}^{t+1},\, x_{L,\mathrm{b}}^t) \\[6pt]
h_{H,\mathrm{b}}^t = \mathrm{DRU}(h_{H,\mathrm{b}}^{t+1},\, x_{H,\mathrm{b}}^t,\, \tilde{h}_{L,\mathrm{b}}^t)
\end{cases}
\end{equation}

At the final time step, the hidden states from both directions are combined to obtain a flow estimate that benefits from past and future evidence, improving robustness against occlusion and acceleration.

\subsection{Anchored STSC Hybrid Training via Curriculum Learning}
\label{sec:CL}


The proposed architecture produces two optical flow predictions, $\{\mathbf{u}^1, \mathbf{u}^2\}$, corresponding to the low-resolution and high-resolution branches. Supervised training is performed using a standard multi-scale endpoint $\ell_1$ loss:

\begin{equation}
\mathcal{L}_{\text{flow}} = \sum_{j=1}^{2} \gamma_j \left\| \mathbf{u}^{gt} - \mathbf{u}^j \right\|_1,
\end{equation}
where $\gamma_j$ is the weight for the $j$-th scale prediction. We set $\gamma_1 = 0.25$ and $\gamma_2 = 0.75$ to emphasize fine-scale accuracy while retaining coarse-level motion guidance.

Although STSC enables fully self-supervised learning of continuous-time flow, direct optimization from scratch is often unstable and prone to collapse.
To provide a reliable initialization anchor, we incorporate sparse ground-truth flow into a curriculum-learning framework. 

During the early training stage, the model relies more on GT to establish stable motion scales and global structure; as training progresses, supervision is gradually reduced and replaced by STSC-based self-supervision, allowing the network to exploit dense temporal consistency cues. The schedule is defined as:


\begin{equation}
\begin{aligned}
    \lambda_{flow}(e)&=\max\left(0, 1-\frac{e}{E_c}\right),\\
    \lambda_{LSC}(e)&=\lambda_{TC}(e)=\frac{1-\lambda_{flow}(e)}{2},
\end{aligned}
\end{equation}
where $e$ denotes the current learning epoch and $E_c$ is the curriculum length. 
The final training objective can be summarized as:

\begin{equation}
    \mathcal{L}=\lambda_{flow}(e)\mathcal{L}_{flow}+\lambda_{LSC}(e)\mathcal{L}_{LSC}+\lambda_{TC}(e)\mathcal{L}_{TC}.
\end{equation}

This anchored hybrid regime follows the curriculum-learning principle:
Training begins with explicit GT supervision and progressively transitions toward predominantly self-supervised learning guided by STSC.

\begin{figure*}[h]
    \centering
    \includegraphics[width=1\linewidth]{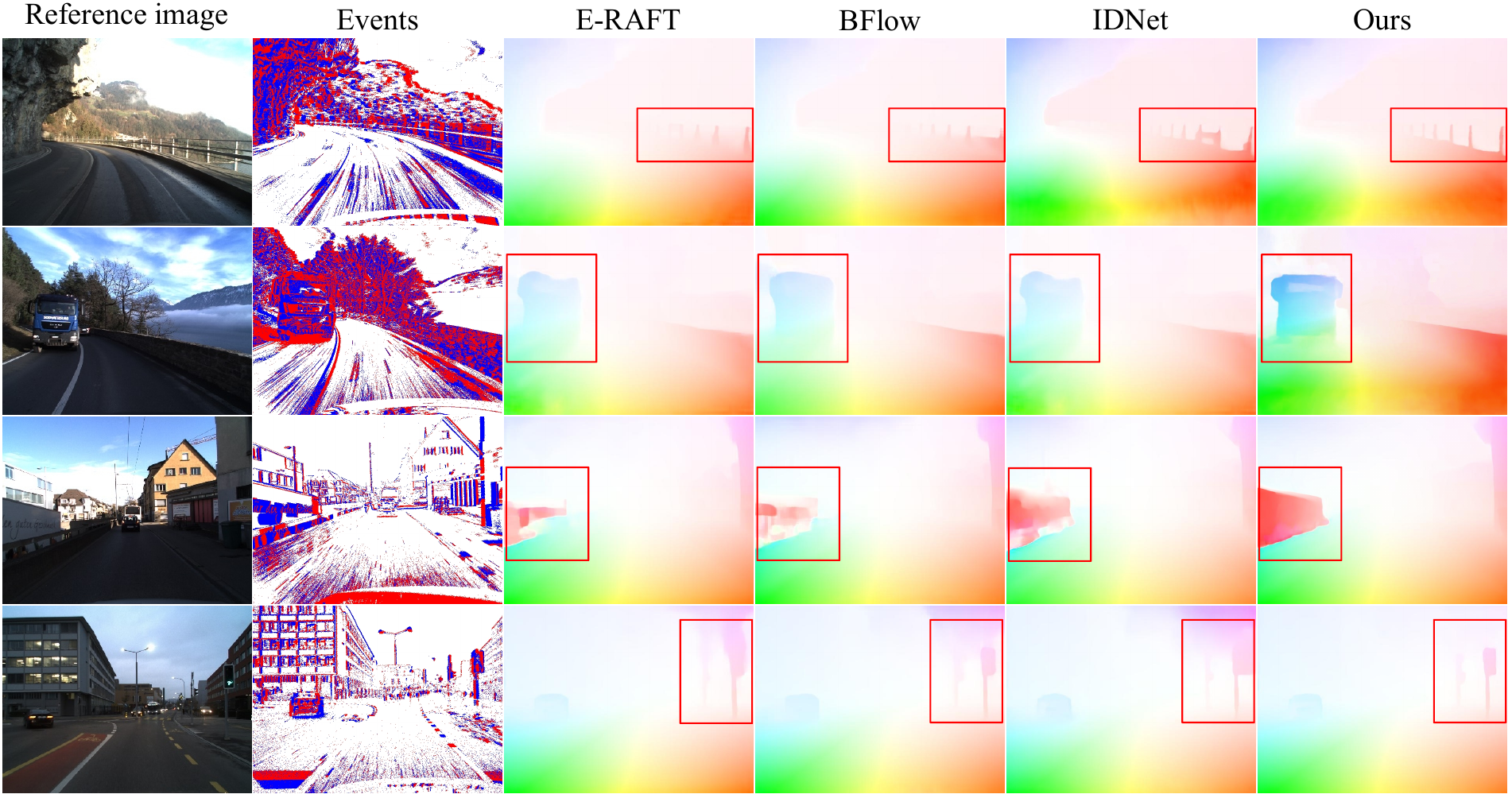}
    \caption{Qualitative results of flow predictions on DSEC-Flow\cite{gehrig2021dsec}. Notable areas are marked by red boxes. Please zoom in for details.}
    \label{fig:dsec}
\end{figure*}

\section{Experiments}

\subsection{Datasets and Implementation Details}

\subsubsection{Datasets.}
We conduct experiments on two widely-used event-based optical flow benchmarks: \textbf{DSEC-Flow}~\cite{gehrig2021dsec} and \textbf{MVSEC}~\cite{zhu2018multivehicle}. 
Our main analysis focuses on DSEC-Flow due to its higher resolution, denser event streams, and more accurate ground-truth flow. 
On DSEC-Flow, the model is trained on the official training split and evaluated on the public test benchmark.
For MVSEC, following prior works, we train on the \textit{outdoor\_day2} sequence and evaluate on \textit{outdoor\_day1}, using two time intervals ($\textit{dt} = 4$ and $\textit{dt} = 1$) to assess performance under different temporal granularities.

\subsubsection{Metrics.}
We evaluate our method using several established metrics. Our primary metric is the \textbf{End-Point Error (EPE)}, defined as the average Euclidean distance between the predicted and ground-truth flow. On DSEC-Flow, we supplement this with \textbf{NPE@1/2/3} (the percentage of pixels with EPE exceeding 1, 2, and 3 pixels) and \textbf{Angular Error (AE)}. On MVSEC, we adhere to the standard protocol by reporting \textbf{\%Out} (the percentage of pixels with EPE exceeding 3 pixels or 5\% of the ground-truth flow magnitude). To specifically assess the accuracy of our continuous high-temporal-resolution (HTR) trajectories, we employ the FWL metric~\cite{stoffregen2020reducing} on the DSEC-Flow dataset. A higher FWL value signifies more accurate HTR flow trajectories.

\subsubsection{Implementation Details.}
Our framework is implemented in PyTorch and trained on a NVIDIA RTX 4090 GPU.
Events within each 100\,ms window are divided into \(B\) temporal bins: \(B{=}15\) on DSEC-Flow; on MVSEC, we use $B=5$ for $\textit{dt}=1$ and $B=15$ for $\textit{dt}=4$. We apply standard augmentations for data diversity: random cropping ($288{\times}384$ for DSEC-Flow, $256{\times}256$ for MVSEC), horizontal flipping (50\%), and vertical flipping (10\%).      
We model continuous motion using quadratic B\'ezier trajectories parameterized by control points.
The model is trained using the Adam optimizer~\cite{kingma2015adam} with a One-Cycle schedule~\cite{smith2019super} and a $1.3 \times 10^{-4}$ peak learning rate. We train for 200 epochs on DSEC-Flow (batch size 2) and 30 epochs on MVSEC (batch size 4). Each sample undergoes four iterative flow updates during training and inference.


\begin{table*}[t]
    \centering
    \caption{
        Evaluation on DSEC-Flow~\cite{gehrig2021dsec}. 
        The best results are in bold, while the second-best ones are underlined. 
        $\downarrow$ means lower is better, $\uparrow$ means higher is better. 
        “HTR” indicates high temporal resolution methods.
    }
    \begin{tabular}{llc@{\hspace{22pt}}c@{\hspace{22pt}}c@{\hspace{22pt}}c@{\hspace{22pt}}c@{\hspace{22pt}}c@{\hspace{22pt}}c@{\hspace{22pt}}}
        \toprule
        & Method & EPE$\downarrow$ & 3PE$\downarrow$ & 2PE$\downarrow$ & 1PE$\downarrow$ & AE$\downarrow$ & FWL$\uparrow$ & HTR \\
        \midrule
        \multirow{1}{*}{MB} & MultiCM~\cite{shiba2022secrets} & 3.472 & 30.86 & 48.48 & 76.57 & 13.98 & 1.37 &  \\
        \midrule
        \multirow{2}{*}{SSL} 
        & TamingCM~\cite{paredes2023taming}    & 2.330 & 17.77 & 33.48 & 68.29 & 10.56 & 1.26 & \checkmark \\
        & Motion-priorCM~\cite{hamann2024motion}    & 3.195 & 15.21 & 25.77 & 53.03 & 8.52 & 1.46 & \checkmark \\
        & E2FAI~\cite{guo2025unsupervised}    & 1.781 & 11.24 & 17.59 & 37.46 & 6.44 & 1.79 &  \\
        \midrule
        \multirow{11}{*}{SL}
        & EV-FlowNet~\cite{zhu2018ev}      & 2.320 & 18.60 & 29.80 & 55.40 & 7.90 & --   &  \\
        & E-RAFT~\cite{gehrig2021raft}     & 0.788 & 2.68  & 4.74  & 12.74 & 2.85 & 1.33 &  \\
        & ADMFlow~\cite{luo2023learning}   & 0.779 & 2.65  & 4.67  & 12.52 & 2.84 & --   &  \\
        & E-FlowFormer~\cite{li2023blinkflow} & 0.759 & 2.45 & 4.10 & 11.23 & 2.68 & -- & \\
        & EEMFlow+~\cite{luo2024efficient} & 0.751 & 2.15  & 3.93  & 11.40 & 2.67 & --   &  \\
        & TMA~\cite{liu2023tma}            & 0.743 & 2.30  & 3.97  & 10.86 & 2.68 & 2.07 &  \\
        & EDCFlow~\cite{liu2025edcflow}    & 0.720 & 2.10  & 3.60  & \underline{10.00} & \underline{2.65} & -- & \\
        & IDNet~\cite{wu2024lightweight}   & \underline{0.719} & \underline{2.04} & \underline{3.50} & 10.07 & 2.72 & 1.97 &  \\
        \cmidrule(lr){2-9}
        & ResFlow~\cite{zhou2025resflow}   & 0.754 & 2.50 & 4.24 & 11.22 & 2.73 & \underline{2.14} & \checkmark \\
        & BFlow~\cite{gehrig2024dense}     & 0.750 & 2.44 & 4.41 & 11.90 & 2.68 & 1.98 & \checkmark \\
        & \textbf{Ours}       & \textbf{0.663} & \textbf{1.60} & \textbf{2.67} & \textbf{7.94} & \textbf{2.53} & \textbf{2.18} & \checkmark \\
        \bottomrule
    \end{tabular}

    \label{tab:dsec}
\end{table*}

\begin{table}[t]
    \centering
    \small
    \setlength{\tabcolsep}{4pt} 
    \caption{
        Evaluation on MVSEC~\cite{zhu2018multivehicle}. 
        The best results are in bold, while the second-best ones are underlined. 
        $\downarrow$ means lower is better.
    }
    \begin{tabular}{llcccc}
        \toprule
        & Method & \multicolumn{2}{c}{$dt=1$} & \multicolumn{2}{c}{$dt=4$} \\
        \cmidrule(lr){3-4} \cmidrule(lr){5-6}
        & & EPE$\downarrow$ & \%Out$\downarrow$ & EPE$\downarrow$ & \%Out$\downarrow$ \\
        \midrule
        MB & MultiCM~\cite{shiba2022secrets} & 0.30 & 0.10 & 1.25 & 9.21 \\
        \midrule
        \multirow{3}{*}{SSL} 
        & EV-FlowNet~\cite{zhu2018ev} & 0.49 & 0.20 & 1.23 & 7.30 \\
        & Spike-FlowNet~\cite{lee2020spike} & 0.49 & -- & 1.09 & -- \\
        & STE-FlowNet~\cite{ding2022spatio} & 0.42 & \textbf{0.00} & 0.99 & 3.90 \\
        \midrule
        \multirow{6}{*}{SL} 
        & E-RAFT~\cite{gehrig2021raft} & 0.27 & \textbf{0.00} & 0.84 & 1.70 \\
        & ADMFlow~\cite{luo2023learning} & 0.52 & \textbf{0.00} & 1.91 & 19.2 \\
        & TMA~\cite{liu2023tma} & 0.25 & \underline{0.07} & 0.70 & 1.08 \\
        & IDNet~\cite{wu2024lightweight} & 0.29 & \textbf{0.00} & 0.75 & 1.20 \\
        & EDCFlow~\cite{liu2025edcflow} & \underline{0.23} & \textbf{0.00} & \underline{0.67} & \underline{0.85} \\
        & \textbf{Ours} & \textbf{0.22} & \textbf{0.00} & \textbf{0.62} & \textbf{0.78} \\
        \bottomrule
    \end{tabular}

    \label{tab:mvsec}
\end{table}

\subsubsection{DSEC-Flow.}
Table~\ref{tab:dsec} reports quantitative results on the DSEC-Flow benchmark, where our method achieves state-of-the-art performance across both standard and HTR-specific metrics.
Our method obtains an EPE of \textbf{0.663}, significantly outperforming all previous methods. We surpass the previous best LTR method, IDNet~\cite{wu2024lightweight} (0.719 EPE), by a relative \textbf{7.8\%}. More importantly, compared to other HTR methods, our EPE is \textbf{11.6\%} lower than BFlow~\cite{gehrig2024dense} (0.750 EPE).

This lead extends across all metrics. We achieve the highest HTR-specific FWL score of \textbf{2.18}, outperforming the previous best HTR method ResFlow~\cite{zhou2025resflow} (2.14). This performance advantage is particularly pronounced in robustness-oriented metrics, which measure errors in challenging regions. Our method achieves SOTA in 3PE (\textbf{1.60}), 2PE (\textbf{2.67}), and 1PE (\textbf{7.94}). These results represent a \textbf{21.6\%} reduction in 3PE and \textbf{23.7\%} in 2PE compared to the prior best LTR model (IDNet). This advantage also holds against HTR methods, showing a \textbf{34.4\%} 3PE reduction and \textbf{39.5\%} 2PE reduction over BFlow, highlighting our method's superior reliability.

Qualitative results visually support these quantitative gains. As shown in Fig.~\ref{fig:dsec}, compared to prior methods~\cite{gehrig2021raft, gehrig2024dense, wu2024lightweight}, our method preserves sharp motion boundaries and resolves fine, spatially varying motion details. Furthermore, Fig.~\ref{fig:occlusion} highlights our robustness near strong motion discontinuities and occluded regions, a benefit largely attributable to the bidirectional accumulation within the BRU module.

\subsubsection{MVSEC.}
As shown in Tab.~\ref{tab:mvsec}, our method achieves SOTA performance on MVSEC across both temporal settings. In the challenging sparse-event condition ($\textit{dt} = 4$), we attain an EPE of \textbf{0.62} and \%Out of \textbf{0.78}, surpassing the prior best EDCFlow~\cite{liu2025edcflow} (0.67 / 0.85) with relative improvements of \textbf{7.5\%} and \textbf{8.2\%}. In the dense setting ($\textit{dt} = 1$), our model achieves the lowest EPE of \textbf{0.22} and matches the best \%Out (0.00). These results confirm our method's robustness across different event densities.


\subsection{Ablation Study}

\begin{figure}[t]
    \centering
    \includegraphics[width=1\linewidth]{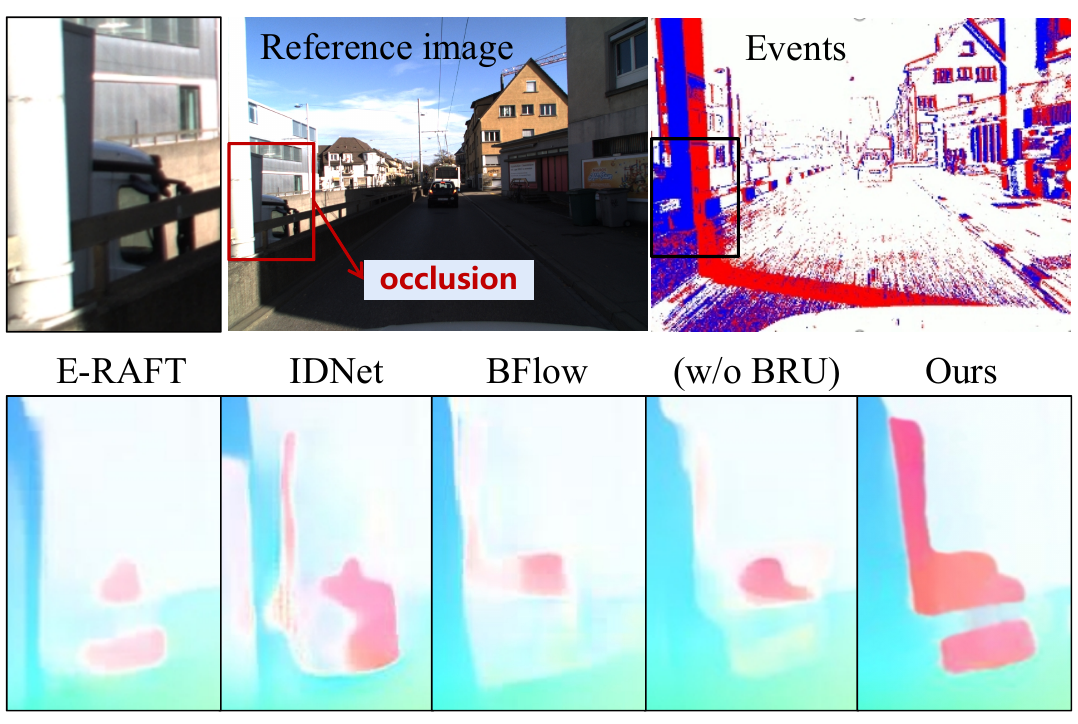}
    \caption{Qualitative comparisons on DSEC-Flow\cite{gehrig2021dsec}. We highlight regions with occlusions and boundaries using red boxes and provide magnified views for better visualization.}
    \label{fig:occlusion}
\end{figure}

\begin{table}[t]
    \centering
    \small
    \caption{Ablation study of architecture components (Bi$^2$ME, SAU, BRU) and our STSC training framework. Best results are in bold.}
    \begin{tabular}{
        c@{\hspace{4pt}}
        c@{\hspace{4pt}}
        c@{\hspace{4pt}}
        c@{\hspace{8pt}}|
        c@{\hspace{10pt}}
        c@{\hspace{10pt}}
        c@{\hspace{10pt}}
        c
    }
        \toprule
        Bi$^2$ME & SAU & BRU & STSC & EPE$\downarrow$ & 3PE$\downarrow$ & 1PE$\downarrow$ & FWL$\uparrow$ \\
        \midrule
         & & & & 0.728 & 2.11 & 10.03 & 1.97 \\
        \midrule
        \checkmark & & & & 0.703 & 1.92 & 9.15 & 1.97 \\
        \checkmark & \checkmark & & & 0.688 & 1.73 & 8.65 & 1.99 \\
        \checkmark & \checkmark & \checkmark & & 0.672 & 1.62 & 8.22 & 2.04 \\
        \checkmark & \checkmark & \checkmark & \checkmark & \textbf{0.663} & \textbf{1.60} & \textbf{7.94} & \textbf{2.18} \\
        \bottomrule
    \end{tabular}

    \label{tab:ablation}
\end{table}

\begin{table}[t]
    \centering
    \caption{Ablation on SAU design. We compare our SAU against two baselines:
    (1) a single-scale GRU on concatenated features, and
    (2) parallel dual-scale GRUs without cross-scale interaction.
    }
    \begin{tabular}{l ccc}
        \toprule
        Method & EPE$\downarrow$ & 3PE$\downarrow$ & 1PE$\downarrow$ \\
        \midrule
        Single-scale (Concat + GRU) & 0.687 & 1.85 & 8.92 \\
        Dual-scale (Parallel GRUs)  & 0.684 & 1.81 & 8.83 \\
        \textbf{Dual-scale (SAU, ours)} & \textbf{0.672} & \textbf{1.62} & \textbf{8.22} \\
        \bottomrule
    \end{tabular}
    \label{tab:ablation_sau}
\end{table}

\begin{table}[t]
\centering
\small
\setlength{\tabcolsep}{8pt} 
\caption{
Ablation study of our STSC component on DSEC-Flow, evaluating HTR trajectory quality. We report the FWL score ($\uparrow$) at varying temporal resolutions. The STSC component consistently improves event alignment, and this improvement becomes more pronounced at higher resolutions.
}
\begin{tabular}{l c c c c}
\toprule
Method & \multicolumn{4}{c}{Temporal Resolution (Hz) (FWL $\uparrow$)} \\
\cmidrule(lr){2-5}
& 10 Hz & 50 Hz & 100 Hz & 150 Hz \\
\midrule
BFlow~\cite{gehrig2024dense} & 2.05 & 2.02 & 2.00 & 1.98 \\
Ours (w/o STSC) & 2.04 & 2.01 & 1.99 & 1.99 \\
\textbf{Ours (STSC)} & 2.07 & 2.14 & 2.17 & \textbf{2.18} \\
\bottomrule
\end{tabular}

\label{tab:htr_quality}
\end{table}

We conduct a series of ablation studies to evaluate the effectiveness of each proposed component. The results are summarized in Tab.~\ref{tab:ablation} on the DSEC-Flow benchmark.

All variants share identical training settings and are based on the IDNet~\cite{wu2024lightweight} backbone, which is a forward-only model operating at $1/4$ spatial resolution.




\subsection{Effect of STSC.}
\begin{figure}[t]
    \centering
    \includegraphics[width=1\linewidth]{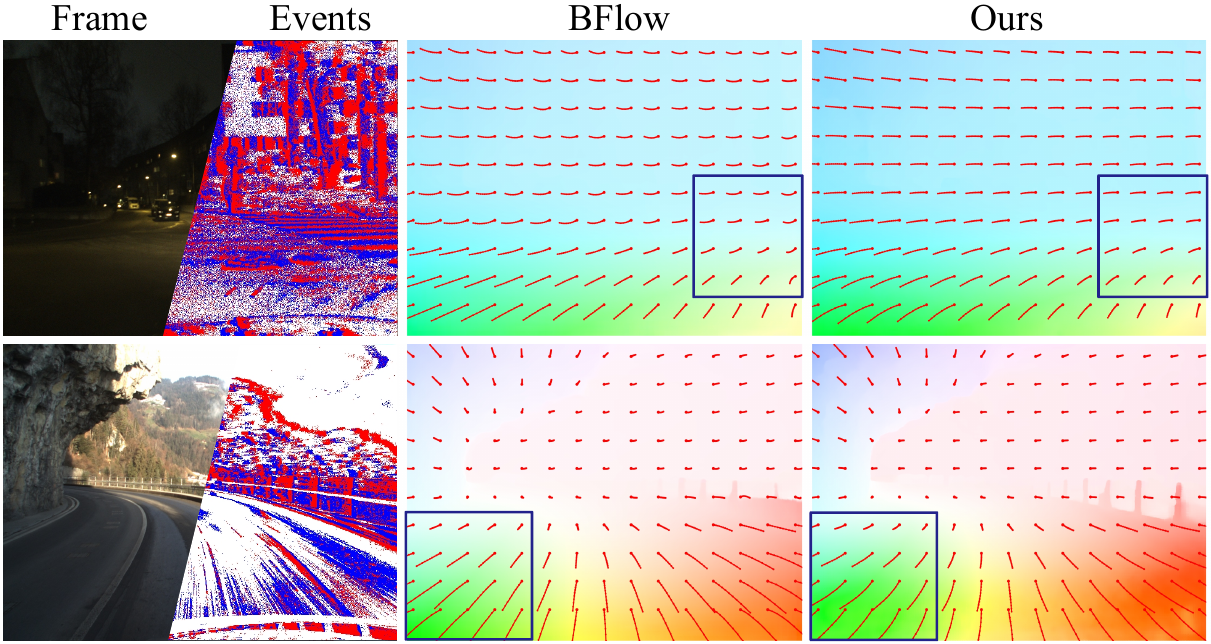}
    \caption{Qualitative visualization in a non-linear motion scene, comparing our method (Ours) to BFlow~\cite{gehrig2024dense}. Blue boxes highlight our method capturing the correct trajectories.}
    \label{fig:traj}
\end{figure}

Table~\ref{tab:htr_quality} further examines the quality of continuous-time trajectories by evaluating the FWL score across different temporal resolutions.
For BFlow~\cite{gehrig2024dense}, the FWL score decreases as resolution increases (2.05 → 1.98), indicating that its trajectories become less physically reliable under denser temporal sampling.
In contrast, our method shows a monotonic improvement (2.07 → 2.18), demonstrating that it produces motion trajectories that remain coherent and physically consistent across time.
This improvement is largely attributed to the enforcement of Spatio-temporal Structural Consistency (STSC): without STSC, FWL quickly degrades with higher sampling rates, whereas incorporating STSC stabilizes trajectory evolution and preserves temporal continuity.
As illustrated in Fig.~\ref{fig:traj}, our approach reconstructs smooth and plausible trajectories rather than merely aligning sparse anchor points, highlighting the advantage of the proposed structure-consistent formulation.

\subsection{Effect of Architecture.}
We then evaluate the components of our network architecture.
Replacing the baseline feature extractor with the Bidirectional Bi-scale Motion Encoder (Bi$^2$ME) yields a clear improvement, reducing EPE by 3.4\% and boosting 3PE by 9\%, confirming its effectiveness in capturing multi-scale bidirectional motion cues.
Building on this foundation, integrating the Scale Alternating Unit (SAU) further enhances performance across all metrics (Tab.~\ref{tab:ablation}). Additional comparisons in Tab.~\ref{tab:ablation_sau} show that SAU consistently outperforms alternatives such as simple bi-scale concatenation or independent single-scale GRUs, underscoring the value of structured cross-scale temporal interaction.
Finally, incorporating the Bidirectional Refinement Update (BRU) module brings further gains by aggregating forward and backward temporal context, leading to improved accuracy and robustness. We also provide qualitative comparisons in Fig.~\ref{fig:occlusion}, visualizing the flow predictions with and without BRU under occlusion and boundary degradation. The model with BRU clearly produces sharper, more coherent motion boundaries, highlighting its enhanced robustness in challenging regions.

\section{Conclusion}
In this paper, we presented a hybrid-supervised framework for continuous-time optical flow estimation, grounded in the principle of spatio-temporal Structural Consistency (STSC). By jointly enforcing local structural stability and trajectory continuity, and further combining a bidirectionally complementary multi-scale architecture with curriculum-guided hybrid training, our method achieves physically coherent motion estimation and establishes new state-of-the-art performance across multiple benchmarks. The proposed STSC formulation suggests a broader theoretical pathway for event-based modeling. Future research may extend this perspective toward unified continuous-time representations, stronger motion priors, and principled regularization of event-driven dynamics.

\section{Acknowledgment}
This work was supported by the Scientific Research Innovation Capability Support Project for Young Faculty (QTZX26002), the National Natural Science Foundation of China (62402366) and the Joint Funds of the Ministry of Education of China (8091B02072401).


{
    \small
    \bibliographystyle{ieeenat_fullname}
    \bibliography{main}
}


\end{document}